# Salient Region Detection and Segmentation in Images using Dynamic Mode Decomposition

Sikha O K[1], Sachin Kumar S[2], K P Soman[2]

[1]Department of Computer Science

[2]Centre for Computational Engineering and Networking

Amrita Viswa Vidyapeetham, Coimbatore - 641 112

**ABSTRACT**

Visual Saliency is the capability of vision system to select distinctive parts of scene and reduce the amount of visual data that need to be processed. The present paper introduces (1) a novel approach to detect salient regions by considering color and luminance based saliency scores using Dynamic Mode Decomposition (DMD), (2) a new interpretation to use DMD approach in static image processing. This approach integrates two data analysis methods: (1) Fourier Transform, (2) Principle Component Analysis. The key idea of our work is to create a color based saliency map. This is based on the observation that salient part of an image usually have distinct colors compared to the remaining portion of the image. We have exploited the power of different color spaces to model the complex and nonlinear behavior of human visual system to generate a color based saliency map. To further improve the effect of final saliency map, we utilized luminance information exploiting the fact that human eye is more sensitive towards brightness than color. The experimental results shows that our method based on DMD theory is effective in comparison with previous state-of-art saliency estimation approaches. The approach presented in this paper is evaluated using ROC curve, F-measure rate, Precision-Recall rate, AUC score etc.

## 1. INTRODUCTION

Human visual system is capable of filtering visually distinctive image regions, rapidly and effortlessly (pre-attentive stage). The filtered part is then processed in detail to obtain high level inferences (attentive stage) and better understanding. The part of an image that stands out from the neighborhood and attracts human attentions in the first sight is termed to be '*salient region'*. Computationally obtaining salient regions has been investigated by cognitive philosophers [2,3] for the last few decades. And was well appreciated by the computer vision community due to its applicability to tackle complex computer vision problems such as object recognition, video compression, automatic image cropping, scene understanding etc. The term saliency was first introduced by Itty et al. in their work on rapid scene analysis in1998 [1]. They come up with a bottom up saliency model capable of capturing spatial discontinuities in scenes. Following Itty's model, a number of methods have been developed in the literature. Zhang et al. proposed an attention model based on fuzzy logic and local contrast analysis [5]. Achanta et al. adopted luminance and color based features to generate saliency maps [6]. In [7], the authors introduced a graph based approach to compute feature based activation map and is then normalized to generate final saliency score. E Rahtu et al. [8] described a statistical framework based saliency detection by utilizing color and motion information. SVD based [9] visual saliency computes saliency map by regularizing singular values of input image. More Recently, Jiwhan et al. used high dimensional features and introduced a trimap based saliency map generation in [10]. In this paper, we have used DMD approach to detect salient regions in static images. To the best of authors knowledge *Static image interpretation of the DMD* method has not been explored in literature till date. Dynamic Mode Decomposition (DMD) is a mathematical theory developed by Peter Schmid in 2008 to simulate and understand complex nonlinear systems without knowing the underlying governing equations of the model [4]. DMD can be used to process experimental data collected in snapshots through time to analyze dimensionality or state of the model and can be used to predict the future state. The theory was originally introduced in natural science and fluid mechanics for time-series data based learning. Recently it was adopted by computer vision community for video processing applications. By nature, video frames are equally spaced in time. The pixels of each snapshot can be vectorized to apply DMD approach which is well suitable for video processing applications such as moving object detection. In [11], authors introduced a real time DMD based method for robustly separating video frames into sparse (foreground) and low-rank (background) components.

## 2. BACKGROUND
### Static Image Interpretation via DMD method

Dynamic Mode Decomposition integrates two of the prominent data analysis methods known today: Fourier Transform and Principle Components. Recently, the theory is adopted to video processing applications as a video sequence is a set of frames that are equally spaced in time and is can be vectorized easily. DMD modes whose Fourier frequencies near to zero (known as zero modes) are considered as background (low-rank component) and those who bounded away from the origin are interpreted as foreground (sparse component). The power of DMD method is still unknown to the static image processing area as there is no interpretation of static images suitable for DMD is available. The work presented in this paper tries to connect Dynamic Mode Decomposition to static image processing applications. The initial objective of the work is to come up with a new interpretation for static images such that it offers an appropriate application for DMD. We have used both color and luminance information to represent a static image as a time-dependent data matrix. To model the nonlinear and highly complex behavior of human visual system, we have exploited the power of different color space representations as each color space provide different color similarity measures.

### Color based representation

In our method, an RGB image of size ($M \times N$) is first converted into YUV, YCbCr and CIELab color spaces. The idea of color space transformation is to have a better model that is capable of separating Luminance (L) from Chrominance(C), since human visual system is more sensitive towards brightness (luminance) than to color information (chrominance) [12]. We used multiple color space representations to provide a series of sequences for DMD algorithm. The Y and L channels from YUV, YCbCr and CIELab are employed to generate salient regions based on luminance and U, V, Cb, Cr, a and b are analyzed based on color information. A color space study has been done prior to the actual experiment to analyze the effect of different color channels on the salient regions. From this, an interesting phenomenon has been observed that salient region of an image has different degree of obviousness in different color channels. A salient region is always prominent in more than one color channel. Figure 1 shows color space study for 'flower pot' and 'ostrich' images. It is obvious that for the 'flower pot' image, the salient region is clear in **a, b, V** and **Cr** channels and for ostrich image **a, U, Cb** color spaces carries more salient information. In figure 1, the image corresponding to the location of bold character denotes prominent salient region image obtained in each color space. Analyzing the color spaces for several images, we came to a conclusion that the combination of **b** from CIELab color space, **V** from YUV color space and **Cr** from YCbCr color space works well for images similar to 'flowerpot' (background and salient part have clear distinction). Similarly, **a** from CIELab color space, **U** from YUV color space and **Cb** from YCbCr color space works well for images similar to the 'ostrich' image (blurred background).

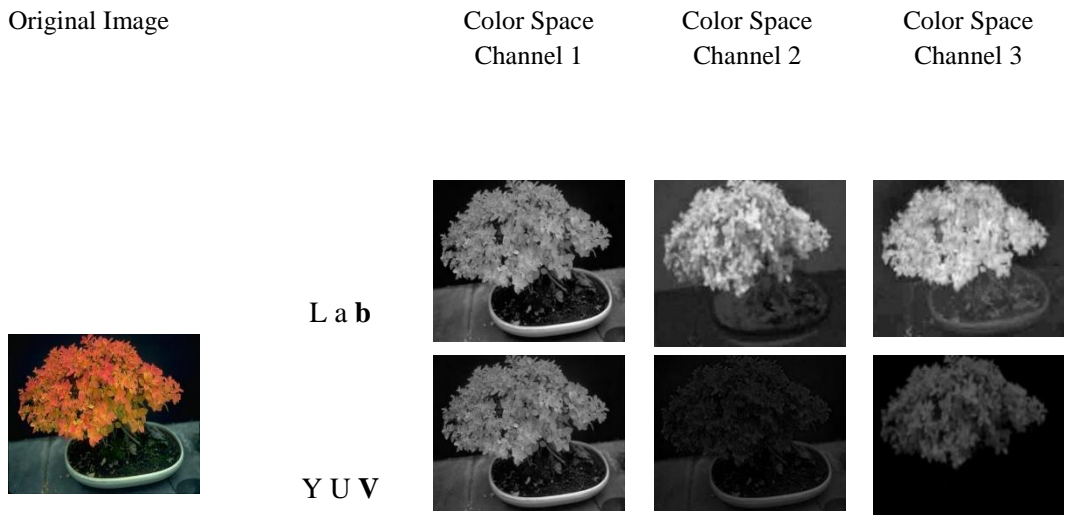

Original Image | Color Space Channel 1 | Color Space Channel 2 | Color Space Channel 3

L a **b**

Y U **V**

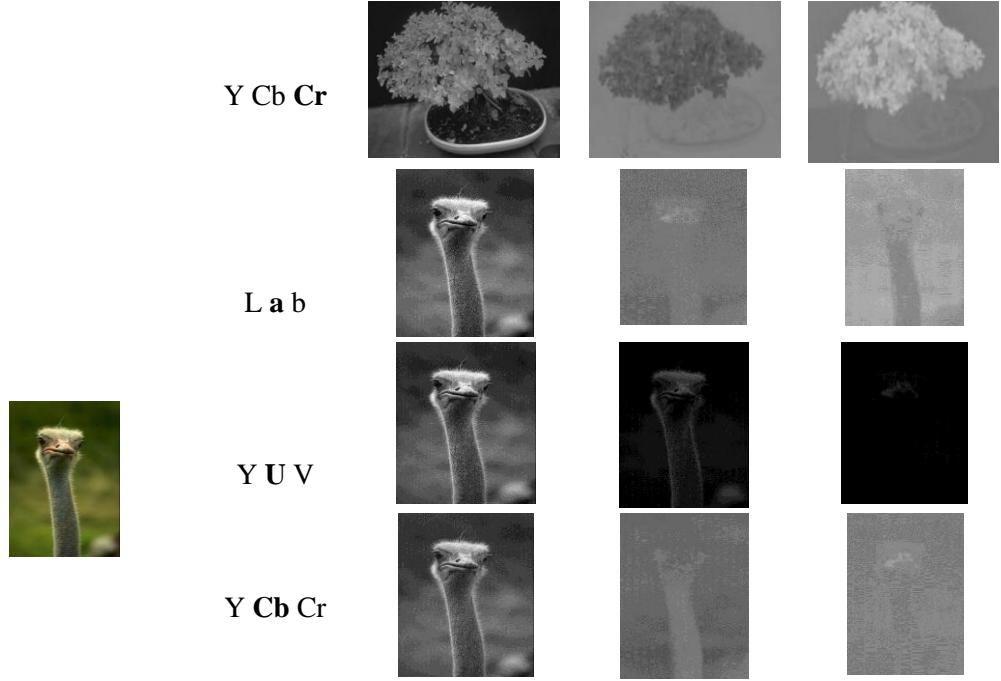

**Figure 1**: Color Space Analysis

Based on the above observations, we formulated two matrices $[C_1]_{mn \times 3}, [C_2]_{mn \times 3}$ which contains color channels denoted as column vectors.

$$C_1 = \begin{bmatrix} | & | & | \\ | & | & | \\ b & V & Cr \\ | & | & | \\ | & | & | \end{bmatrix} \quad C_2 = \begin{bmatrix} | & | & | \\ | & | & | \\ a & U & Cb \\ | & | & | \\ | & | & | \end{bmatrix}$$

The columns of matrices $[C_1]_{mn \times 3}, [C_2]_{mn \times 3}$ are permuted and repeated to form an input for DMD method to generate color based saliency. Repetition of columns are suggested as DMD method gives accurate separation of salient part from the background if the number of columns of the underlying data matrix is as large as possible which has been proved experimentally[13].

**Luminance based representation**

In order to incorporate the luminance information into the DMD approach, we have used singular values computed using SVD (Singular Value Decomposition) of luminance components (L, Y). . We adopt the hypothesis introduced by Xiaolong et al. in [9] that intermediate singular values represents salient part of an image while smaller and larger values represents non salient parts. Figure 2 shows the reconstructed SVD images for different set of singular values based on 'L' color channel of CIELab color space. In Figure 2, (a) shows the original image (b) shows image reconstructed using the largest singular value which contains the background information. Figure 2 (d)-(f) shows images reconstructed using respective combinations of singular values shown at the bottom of the figure. It is evident that starting from figure 2(b) each of the reconstructed image contains more information about the salient object. As the number of singular values used for reconstruction increases, salient region become more clear while the background remains relatively static after a set of iterations. Based on the above observation it is noticed that SVD reconstructed images for various intermediate singular values offers an appropriate application for DMD method.

Each of the reconstructed SVD images are reshaped to 1-dimensional column vectors and are united into single data matrix $I$. Given $\ell$ intermediate singular values, the $n \times 1$ vectors $x_1, x_2, x_3.....x_\ell$ are extracted, where each $x_i$ represents vectorized SVD image. The obtained time-dependent matrix is then fed into DMD algorithm for low rank-sparse separation.

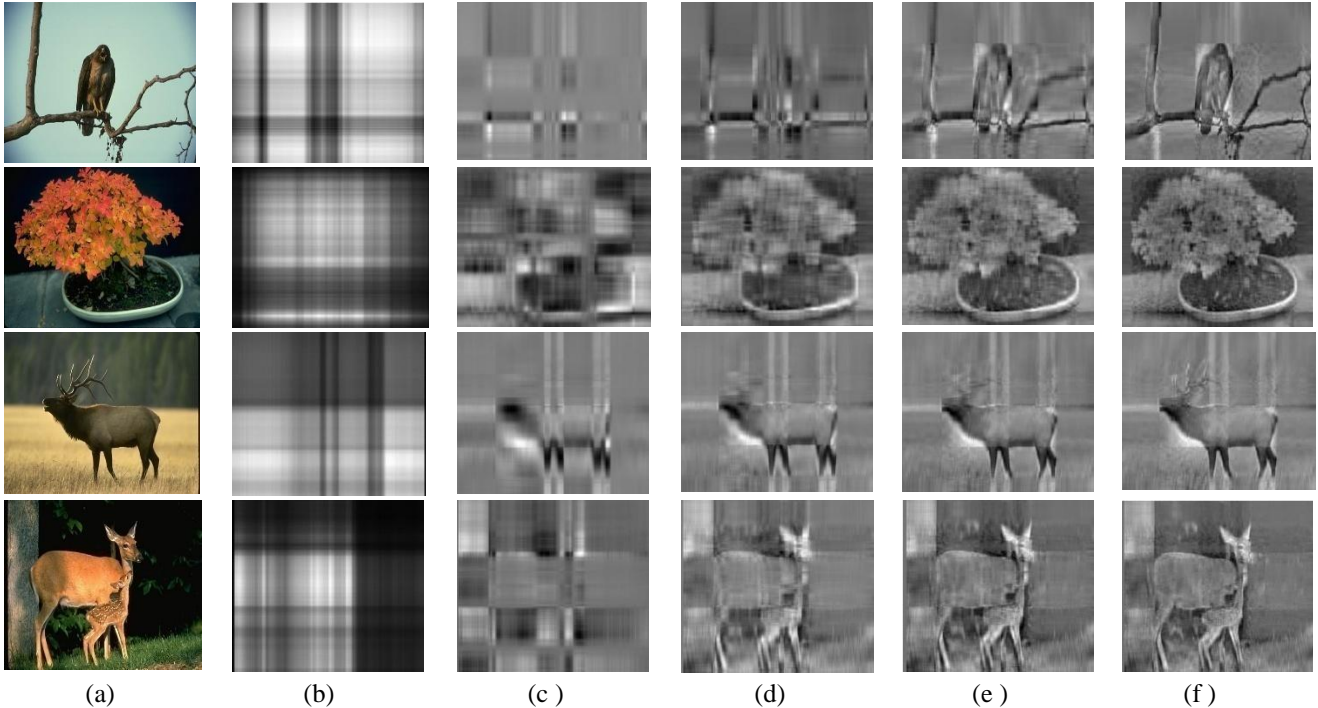

(a) (b) (c) (d) (e) (f)

**Figure 2**. SVD reconstructed image for various singular values (a) Original image (b) Largest singular value (c) Singular values 3-5 (d) Singular values 3-10 (e) Singular values 3-15 (f) Singular values (3-20)

## 3. RESULTS OBTAINED

The dataset used for the evaluations are taken from [14] along with few downloaded natural scene images from several web pages. Obtained results are shown in figure 3.

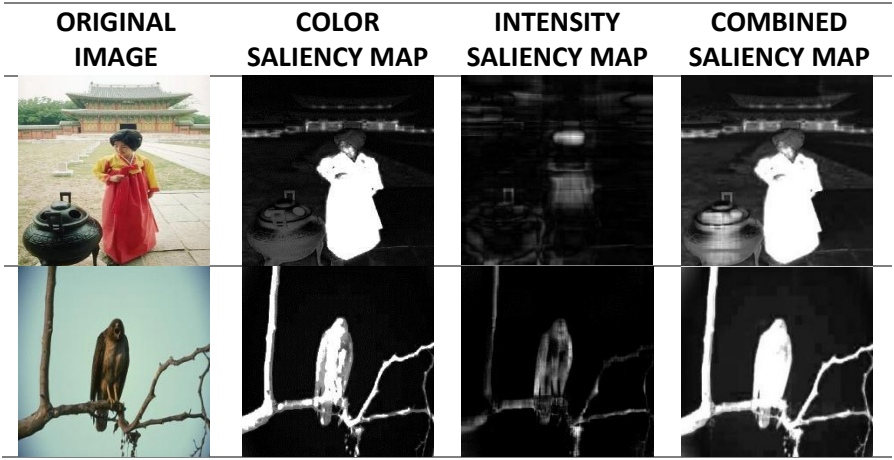

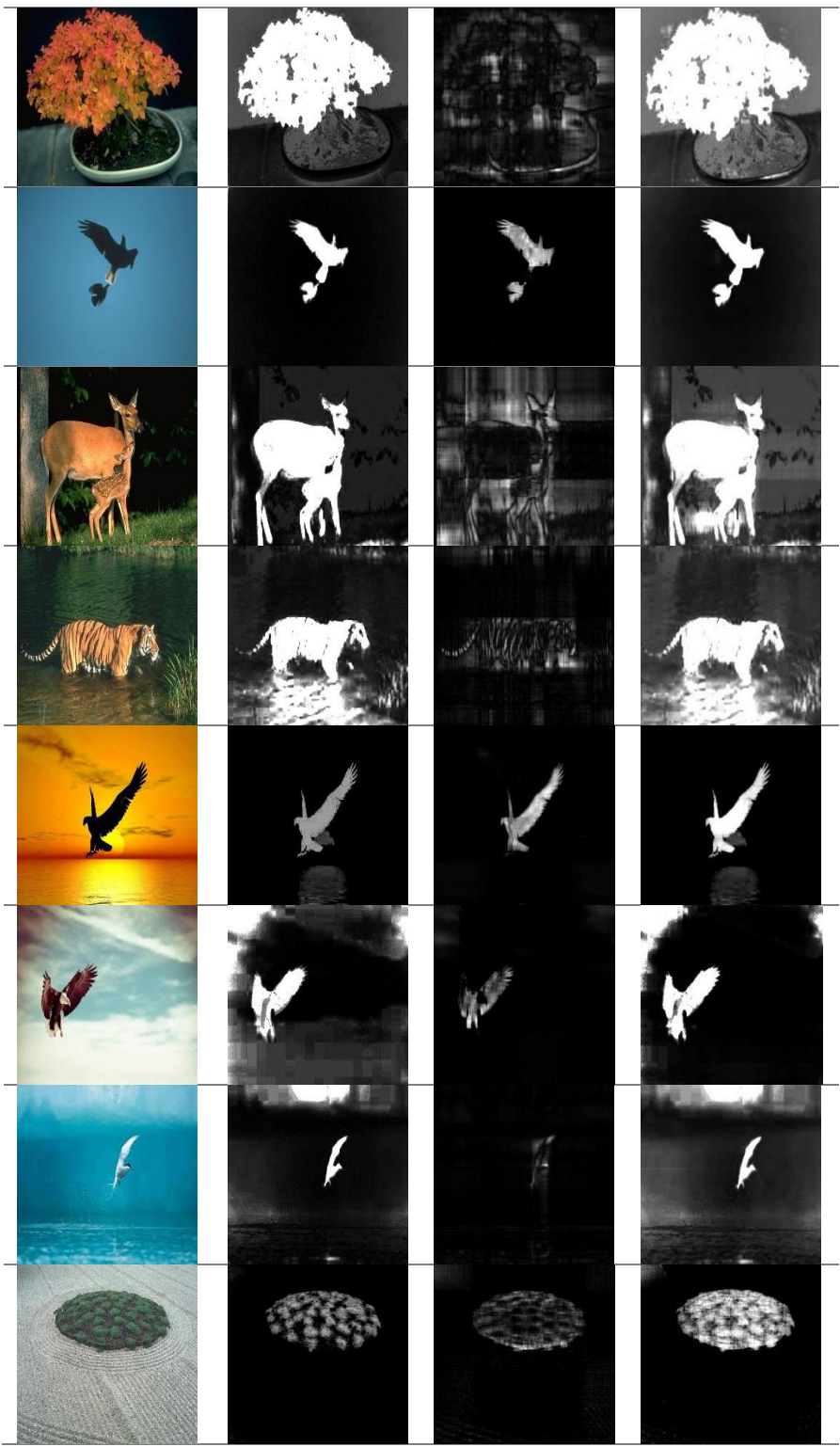

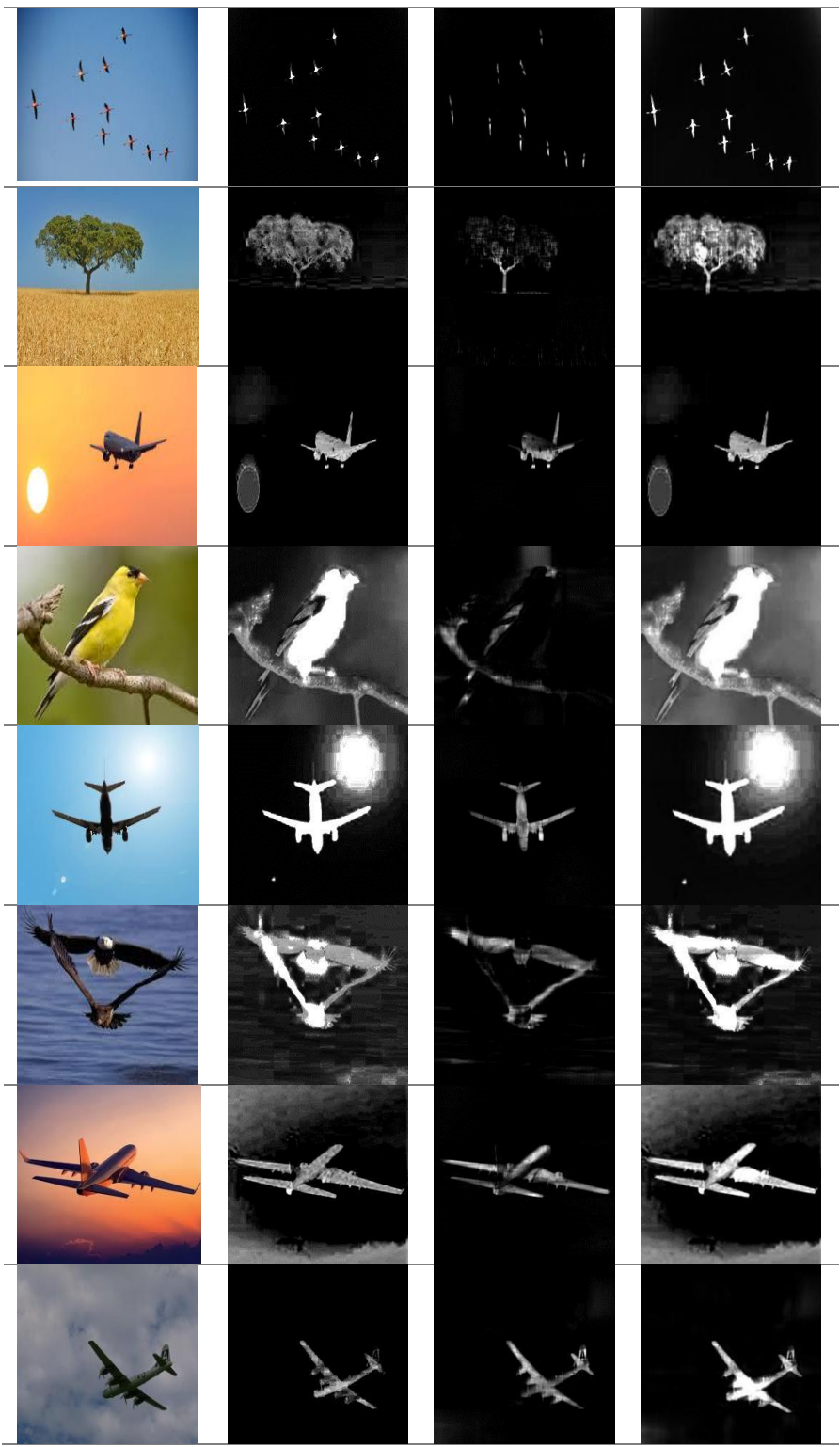

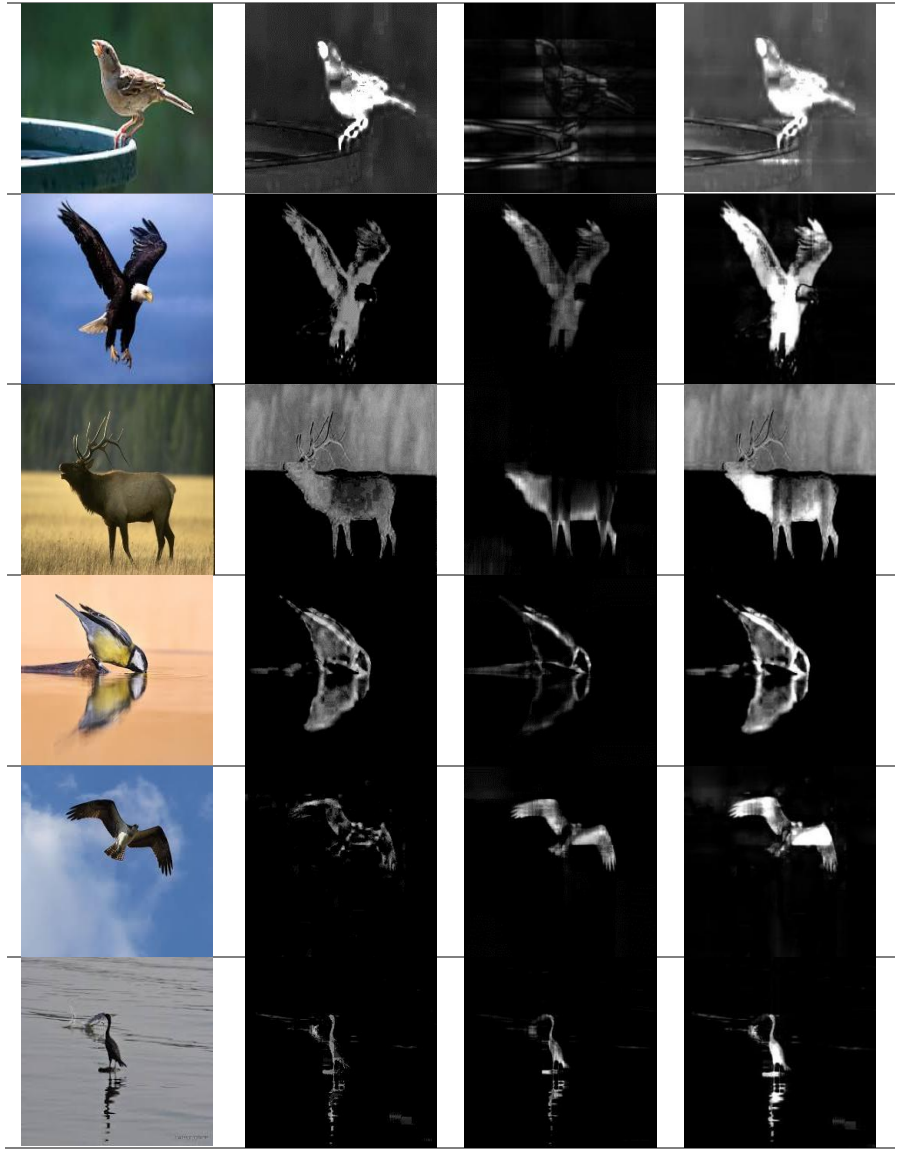

**Figure 3.** Results

Salient objects are segmented out from the obtained saliency map, and the results are evaluated against state-of-art saliency detection algorithms in terms of ROC curve, F-measure rate, Precision-Recall rate, AUC score etc. The details of the evaluations with entire work will be published soon.

4. **CONCLUSION**

We have presented a Dynamic Mode Decomposition based salient region detection. Saliency detection is considered as a problem of matrix separation into low-rank and sparse. The paper also introduces a new interpretation to use DMD approach in static image processing. This approach integrates two data analysis methods: (1) Fourier Transform, (2) Principle Component Analysis. The experimental results shows that our method based on DMD theory is effective in comparison with previous state-of-art saliency estimation approaches. The approach presented in this paper is evaluated using ROC curve, F-measure rate, Precision-Recall rate, AUC score etc.